\documentclass{article}
\usepackage[final, nonatbib]{nips_2016} 

\usepackage[utf8]{inputenc}
\usepackage{algpseudocode, algorithm}
\usepackage{graphicx}
\usepackage{hyperref}
\usepackage{subcaption}
\usepackage{amssymb}

\title{An End-to-End Architecture for Keyword Spotting and Voice Activity Detection}
\author{\textbf{Chris Lengerich}\thanks{Equal contribution} \\Mindori\\Palo Alto, CA\\ \texttt{chris@mindori.com} \and \textbf{Awni Hannun}\footnotemark[1]\\Mindori\\Palo Alto, CA\\\texttt{awni@mindori.com}}
\date{\{chris, awni\}@mindori.com}

\begin{document}

\maketitle

\begin{abstract}
We propose a single neural network architecture for two tasks: on-line keyword spotting and voice activity detection. We develop novel inference algorithms for an end-to-end Recurrent Neural Network trained with the Connectionist Temporal Classification loss function which allow our model to achieve high accuracy on both keyword spotting and voice activity detection without retraining. In contrast to prior voice activity detection models, our architecture does not require aligned training data and uses the same parameters as the keyword spotting model. This allows us to deploy a high quality voice activity detector with no additional memory or maintenance requirements.
\end{abstract}

\section{Introduction}

Keyword spotting (KWS) is a speech task which requires detecting a specific word in an audio  signal, commonly for use as the ``wake word" of a large-vocabulary (LV) speech recognizer. Voice activity detection (VAD) requires detecting human speech in the signal, often for the purpose of endpointing in a large vocabulary speech recognition system.   Both tasks are challenging due to computational constraints and noisy environments. To limit computational cost, VAD models have often leaned on hand-engineered features, requiring training  separately from KWS models.

We propose instead a single end-to-end neural network architecture for both KWS and VAD. We develop novel inference algorithms which allow us to run  KWS and VAD tasks without retraining. Our model outperforms both  baselines, is trained only on  unaligned character-level transcripts, and requires maintaining only a single architecture for training and deployment.

\section{Related Work}
\label{section:related}

The model is based on work in end-to-end speech recognition which uses the Connectionist Temporal Classification loss function coupled with deep Recurrent Neural Networks \cite{Graves2014, Hannun2014deepspeech}.   In this work we develop the model and inference procedure for the KWS and VAD tasks.  A thorough treatment of the benefits of this model for LVCSR is given in \cite{Amodei2016}. A character-level CTC architecture was also recently adopted for keyword spotting \cite{Hwang2015}, where it outperformed a DNN-HMM baseline, while a word-level CTC architecture was used for keyword spotting in \cite{Fernandez2007}.

Traditional VAD architectures trade off accuracy for low computational cost, as they have historically been developed for very low-resource environments. Some simple and efficient techniques include a threshold on the energy of the audio signal, a threshold on the number of zero-crossings \cite{Junqua1991} or combinations of these features, however, these methods are typically not robust to non-stationary environments. We note that modern LV speech environments afford more computational resources than before, and the use of larger neural models is feasible, especially for LV ASR endpointing. Neural architectures have been proposed for VAD, notably the RNN architecture in \cite{Hughes2013vad}, however that approach relied on frame-aligned labels.

\section{Model}
\label{section:model}

For a general keyword spotter, we model $p(k | x)$, where $k$ is a keyword and $x$ is a window of speech. For VAD we use the same distribution and simply set $k$ to the empty string.

We use the Connectionist Temporal Classification \cite{Graves2006} (CTC) objective function to train an RNN on a corpus of utterance and transcription pairs. The CTC objective gives us the probability of any label string for a given utterance. We do not need an alignment as CTC efficiently computes the score over all possible alignments. The objective function for an utterance $x$ and corresponding transcription $\ell$ is given by
\label{alg:ctc}
\begin{equation}
p_{\mathrm{CTC}}(\ell | x) = \sum_{s \in \mathrm{align}(\ell, T)} \prod_t^T p(s_t | x).
\end{equation}
The $\mathrm{align}(\cdot)$ function computes the set of possible alignments of the transcription $\ell$ over the $T$ time-steps of the utterance under the CTC operator. The CTC operator allows for repetitions of any character and insertions of the blank character, $\epsilon$, which signifies no output at a given time-step.

For the on-line KWS task we must determine with low latency if the keyword has been said. In order to use this model for KWS, we score a moving window of the audio stream $x$ so that we can find the keyword soon after it occurs. The score is computed as $p_{\mathrm{CTC}}(k | x_{t:t+w})$ where $k$ is any keyword and $x_{t:t+w}$ is a window of speech $w$ frames long. For the VAD task we first compute the probability of no speech by setting $k$ to the empty string. From this we can find the probability of speech by taking one minus the probability of no speech.

\subsection{Network Architecture}
\label{section:architecture}

The network accepts as input a spectrogram computed from the raw waveform sampled at 8kHz. The first layer is a 2-dimensional convolution with a stride of three \cite{Amodei2016}. For the next three layers of the network we use gated recurrent RNN layers \cite{Cho2014}\cite{Chung2014}. The last layer is a single affine transformation followed by a softmax. The network outputs directly to characters in the alphabet including the blank and space characters.

\subsection{Inference}

\begin{algorithm}
\caption{Compute the score of a keyword, $k$, given the CTC output probabilities, $P$. The parameter $l$ is the keyword with $\epsilon$ inserted at the beginning, end and between every pair of characters of $k$ .}
\label{alg:scorekeyword}
\begin{algorithmic}
\Function{ScoreKeyword}{$l$, $P$}
\State $S$ $\gets$ size($l$)
\State $T$ $\gets$ numberOfColumns($P$)
\State $\alpha \gets$ zeros(S, T)
\State $\alpha_{1, 1} \gets 1 - P_{l_2, 1}$
\State $\alpha_{2, 1} \gets P_{l_{2}, 1}$
    \For{$t=2:T$}
        \For{$s=1:S$}
            \If{$s = 1$}
                \State $p \gets 1 - P_{l_2, t}$
            \ElsIf{$s = S$}
                \State $p \gets 1 - P_{l_{S - 1}, t}$
            \Else
                \State $p \gets P_{l_{s}, t}$
            \EndIf
            
            \If{$s > 2$ and $l_s \ne l_{s-2}$ and $l_s \ne \epsilon$}
                \State $\alpha_{s, t} \gets p * (\alpha_{s, t-1} + \alpha_{s-1, t-1} + \alpha_{s-2, t-1})$
            \ElsIf {$s > 1$}
                \State $\alpha_{s, t} \gets p * (\alpha_{s, t-1} + \alpha_{s-1, t-1})$
            \Else
                \State $\alpha_{s, t} \gets p * \alpha_{s, t-1}$
            \EndIf
        \EndFor
    \EndFor
    \State \Return $\alpha_{S, T} + \alpha_{S-1, T}$
\EndFunction
\end{algorithmic}
\end{algorithm}

The optimal window size for KWS detection varies per utterance based on speech rate, noise and adjacent utterances. To alleviate the sensitivity of the algorithm to the window size parameter we propose a modification to the CTC scoring algorithm presented above. For a given keyword $k$ instead of scoring $k$ itself under the model we instead score the regular expression [\^{}$k_0$]*$k$[\^{}$k_{n-1}$]*, where $k_0$ and $k_{n-1}$ are the first and last characters of $k$, respectively. This is described in  Algorithm \ref{alg:scorekeyword}.

Computing the VAD score reduces to summing the log probabilities of the blank character over the window of speech frames:
\begin{equation}
\log p(\mathrm{speech}| x_{t:t+w}) = 1 - \sum_{i=t}^{t + w} \log p_i(\epsilon | x_{t:t+w}).
\end{equation}

\section{Experiments}
\label{section:experiments}

The model parameters are optimized with stochastic gradient descent for 50 epochs and a minibatch size of 256.  We sort examples so that the minibatch consists of utterances of similar length for computational efficiency. The learning rate and momentum parameters are chosen to optimize speed of convergence. We anneal the learning rate by a factor of 0.9 every 5000 iterations.

The architecture of the network is as described in Section \ref{section:architecture}. The filters for the convolution layer are 11 by 32 over the time and frequency dimensions respectively. We use 32 filters in all models.

\subsection{Data}
The data used to train the model consists of two datasets. The first dataset is a corpus of 526K transcribed utterances collected on Android phones via an assistant-like application. The second corpus consists of 1544 spoken examples of the keyword, in this case, ``Olivia''. The model is trained on both data-sets simultaneously. We do not need to pre-train on the large corpus prior to fine-tuning. We also use a collection of about a hundred hours of noise and music downloaded from the web to generate synthetic noisy examples of the keyword and empty noise clips. When training with the noisy data we replicate each keyword 10 times, each time with a random noise clip. We also use a corpus of 57K randomly sampled noise clips with a blank label as filler.

The KWS model is evaluated on a test set of 550 positive examples (e.g. containing the keyword ``Olivia``) and 5000 negative examples held-out from the large speech corpus described above. During inference we evaluate the utterance with Algorithm \ref{alg:scorekeyword} every 100 milliseconds over a window of 800 milliseconds in order to detect the presence of the keyword. We classify an example as positive if the score found from the output of Algorithm \ref{alg:scorekeyword} over the utterance is ever above a preset threshold.

We evaluate the same models on the VAD task. The positive examples are the same 5000 examples of speech used as the negative examples for the KWS task. We collected about 10 hours of non-speech audio from a variety of noise backgrounds. We sample 5000 random clips from the 10 hours of noise to construct the negative samples.

\subsection{Results}

Our KWS baseline is a DNN keyword spotter from kitt.ai \footnote{\url{https://github.com/Kitt-AI/snowboy}}. Our VAD baseline is the WebRTC VAD codec\footnote{\url{https://github.com/wiseman/py-webrtcvad}} with frame size 30ms. Our model of 3 layers of size 256  outperforms both  baselines. For a fixed false positive rate of 5\% our model achieves a 98.1\% true positive rate on keyword spotting, in comparison to the baseline 96.2\%. For the VAD task, for the same false positive rate, our model achieves 99.8\% true positive rate vs. 44.6\% for the baseline. This large delta may be due to the substantial  difference in representational power of a large-parameter neural model vs. the small-parameter GMM baseline, as well as differences in the type and volume of  training data.

\begin{figure}
\centering
\begin{subfigure}{0.49\textwidth}
    \includegraphics[width=\textwidth]{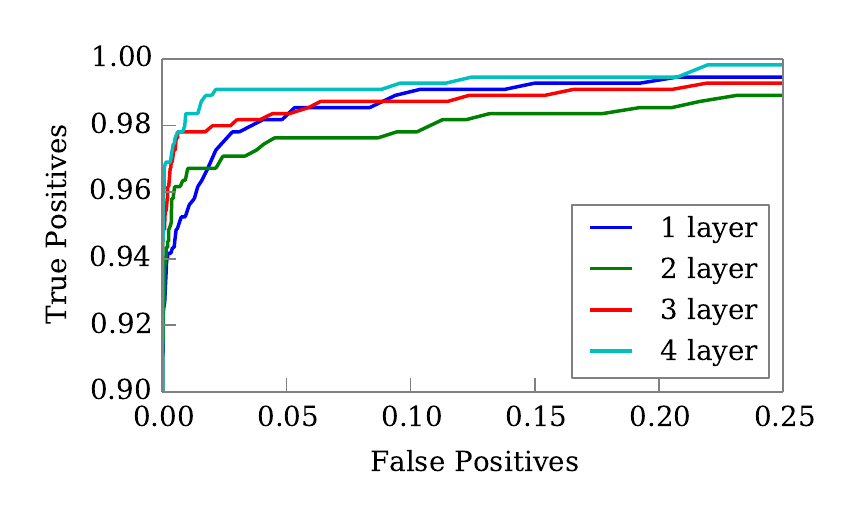}
    \caption{KWS}
\end{subfigure}
\hfill
\begin{subfigure}{0.49\textwidth}
    \includegraphics[width=\textwidth]{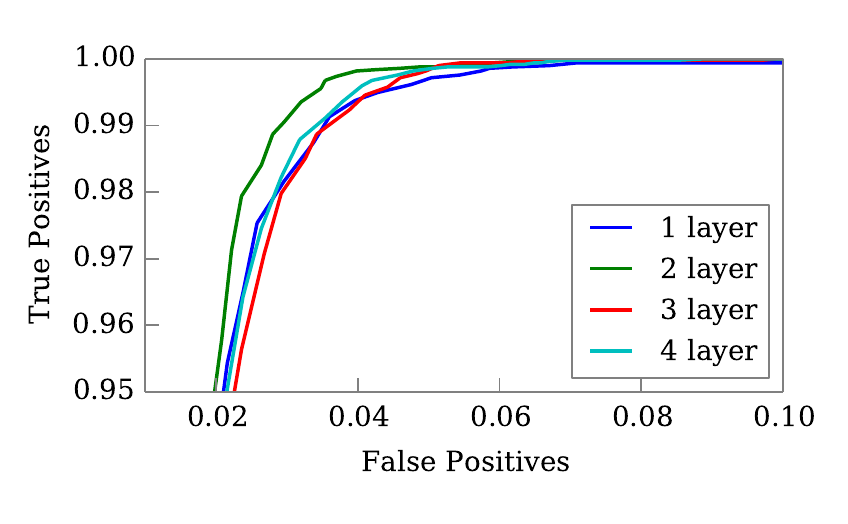}
    \caption{VAD}
\end{subfigure}
\caption{Increasing the number of hidden layers improves performance in both tasks until saturation. The layer size for all models is fixed at 256 hidden units.}
\label{fig:layers} 
\end{figure}

\begin{figure}
\centering
\begin{subfigure}{0.49\textwidth}
    \includegraphics[width=\textwidth]{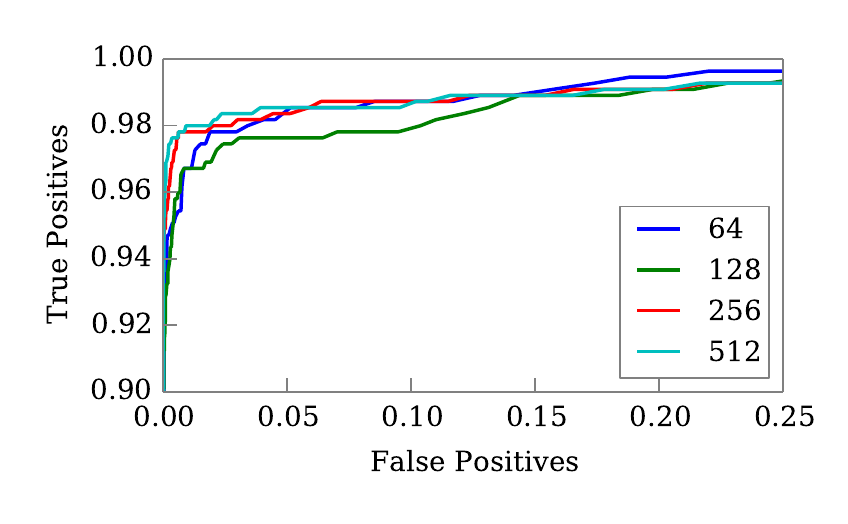}
    \caption{KWS}
\end{subfigure}
\hfill
\begin{subfigure}{0.49\textwidth}
    \includegraphics[width=\textwidth]{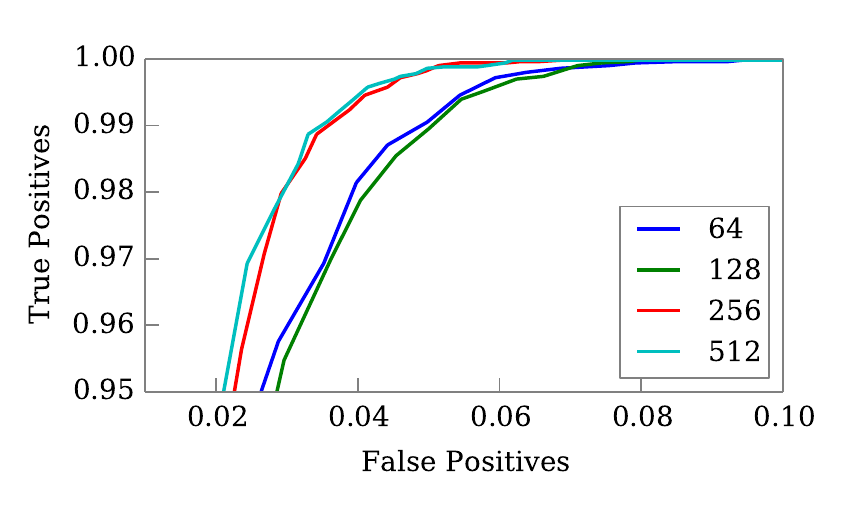}
    \caption{VAD}
\end{subfigure}
\caption{Increasing the layer size also increases performance. The number of hidden layers for all models is fixed at 3.}
\label{fig:sizes} 
\end{figure}

Figures \ref{fig:layers} and \ref{fig:sizes} show that our  model consistently improves at detecting the keyword as we increase the number of layers and the size of the model. Increasing the model depth and size also improves VAD performance; however, when the layers are larger than 128 units or there are more than 2 layers, VAD performance saturates. Most of the large VAD models achieve near 99.9\% true positive rate or higher at a fixed false positive rate of 5.0\%.

\begin{figure}
\centering
\begin{subfigure}{0.49\textwidth}
    \includegraphics[width=\textwidth]{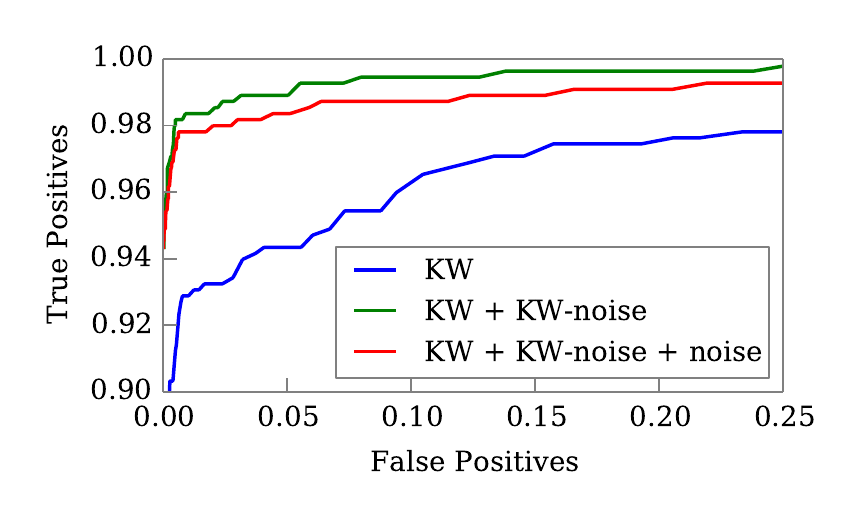}
    \caption{KWS}
\end{subfigure}
\hfill
\begin{subfigure}{0.49\textwidth}
    \includegraphics[width=\textwidth]{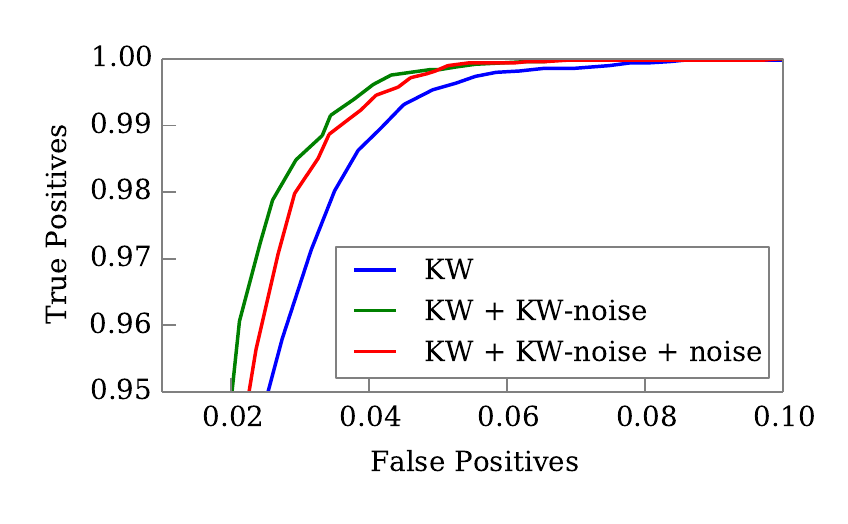}
    \caption{VAD}
\end{subfigure}
\caption{Adding noisy training data improves KWS and VAD performance. The `KW' model contains the 550K filler examples without noise synthesis. The `KW + KW noise` model also includes the keyword data replicated 10 times with random noise. The `KW + KW noise + noise` also includes 57K random noise clips as filler.}
\label{fig:data} 
\end{figure}

In Figure \ref{fig:data} we see that adding noise to the keywords during training results in substantial improvements. At a false positive rate of 5\% the model with noise has a true positive rate of 98.9\% compared to 94.3\% for the model without noise. Further using the random noise data on its own does not help much; in fact the results are slightly worse. On the VAD task we also notice an improvement in the ROC curve as we add noise.

Our production model with 3 layers of 256 hidden units has  $\sim$1.5M trainable parameters, comparable to other neural network-based KWS approaches \cite{Chen2014kws}, and has been deployed to a modern smartphone.

\section{Conclusion}
\label{section:conclusion}
We have described a single neural network architecture which can be used for both keyword spotting and voice activity detection without retraining. The model is simple to train and does not require an alignment or frame-wise labels, in contrast to prior VAD models. We propose inference algorithms for KWS and VAD modified from the basic CTC scoring algorithm which allow the model to perform both tasks.  While our model is efficient, applying neural compression techniques might further increase performance and represents an interesting area for future work.

\end{document}